\documentclass[runningheads]{llncs}
\usepackage[T1]{fontenc}
\usepackage{graphicx}
\usepackage{siunitx}   
\usepackage{booktabs}  
\usepackage{multirow} 
\usepackage{xcolor}
\usepackage{float} 
\usepackage[table]{xcolor}
\usepackage{placeins}

\usepackage{marvosym} 

\begin{document}
\title{GS-DMSR: Dynamic Sensitive Multi-scale
Manifold Enhancement for Accelerated
High-Quality 3D Gaussian Splatting}
%
%
\author{Nengbo Lu \inst{1} \and
Minghua Pan  \textsuperscript{(\Letter)} \inst{2,3} \and 
Shaohua Sun\inst{1} \and
Yizhou Liang\inst{1}}
\authorrunning{F. Author et al.}
%
\institute{School of Artificial Intelligence, Guilin University of Electronic Technology, Guilin,
Guangxi, 541004, China \and
School of Computer Science and Information Security, Guilin University of
Electronic Technology, Guilin 541004, China\\
\email{panmh@guet.edu.cn}\\
\and
Guangxi Key Laboratory of Cryptography and Information Security, Guilin University of Electronic Technology, Guilin 541004, China\\
}
\authorrunning{F. Author et al.}
%
\maketitle              
\begin{abstract}
In the field of 3D dynamic scene reconstruction, how to balance model convergence rate and rendering quality has long been a critical challenge that urgently needs to be addressed, particularly in high-precision modeling of scenes with complex dynamic motions. To tackle this issue, this study proposes the GS-DMSR method. By quantitatively analyzing the dynamic evolution process of Gaussian attributes, this mechanism achieves adaptive gradient focusing, enabling it to dynamically identify significant differences in the motion states of Gaussian models. It then applies differentiated optimization strategies to Gaussian models with varying degrees of significance, thereby significantly improving the model convergence rate. Additionally, this research integrates a multi-scale manifold enhancement module, which leverages the collaborative optimization of an implicit nonlinear decoder and an explicit deformation field to enhance the modeling efficiency for complex deformation scenes. Experimental results demonstrate that this method achieves a frame rate of up to 96 FPS on synthetic datasets, while effectively reducing both storage overhead and training time.Our code and data are available at https://anonymous.4open.science/r/GS-DMSR-2212.

\end{abstract}

\section{Introduction}



Novel view synthesis, as a core research direction in 3D vision, plays a pivotal role in industries such as virtual reality, augmented reality, and film production. This technology aims to construct spatio-temporally continuous representations of scenes from sparse 2D image inputs, enabling dynamic rendering from arbitrary viewpoints and timestamps. Particularly in dynamic scene modeling, the precise reconstruction of complex motion patterns from spatio-temporally limited input data remains a critical challenge in current research.

Neural Radiance Fields (NeRF)\cite{mildenhall2021nerf} have achieved groundbreaking progress in novel view synthesis by representing scenes through implicit functions. This method employs volume rendering techniques to establish mappings between 2D images and 3D scenes. However, the original NeRF suffers from inefficiencies in training and rendering. Although subsequent improvements have reduced training times from days to minutes, its rendering process still incurs significant latency, falling short of real-time requirements.

3D Gaussian Splatting (3D-GS)\cite{kerbl20233d}, an innovative approach using explicit scene representation, marks a major breakthrough in 3D reconstruction. By modeling scenes with explicit 3D Gaussian distributions, 3D-GS elevates rendering speeds to real-time levels. Unlike the computationally intensive volume rendering in NeRF, 3D-GS introduces differentiable splatting techniques to directly project 3D Gaussians onto 2D imaging planes. This representation not only achieves real-time rendering but also provides an explicit scene structure, facilitating scene manipulation and editing, thereby expanding possibilities for scene reconstruction.

\begin{figure*}
    \centering
    \includegraphics[width=1\linewidth]{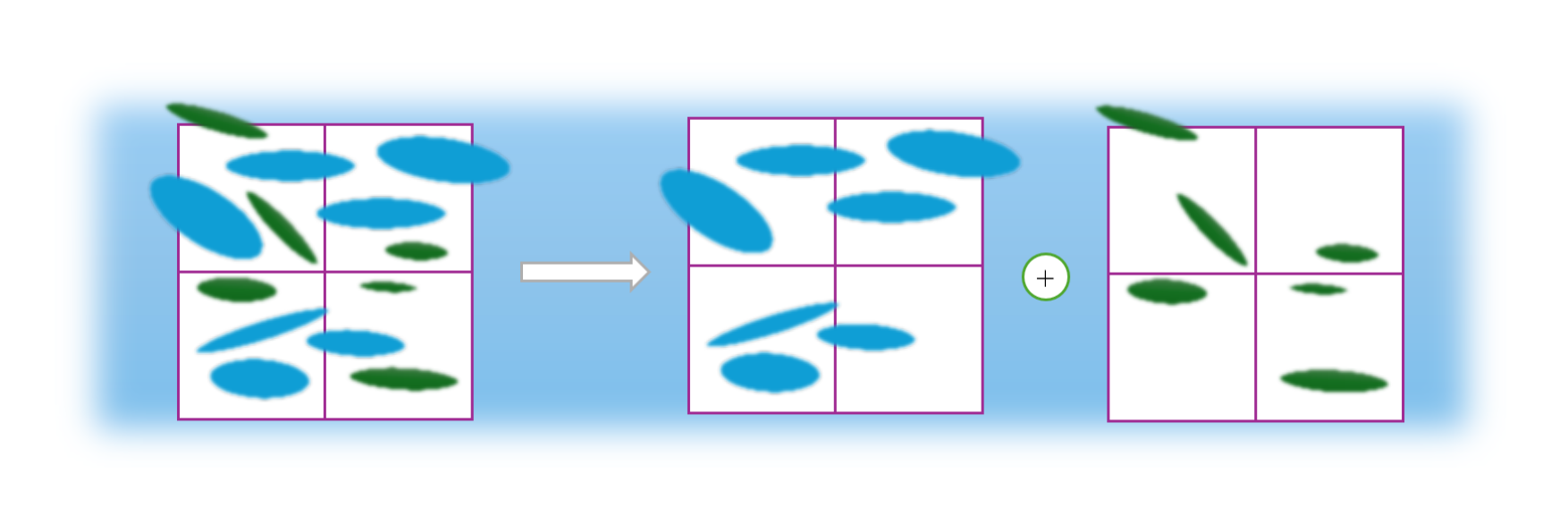}
    \caption{Motion Saliency-Driven Dynamic Gaussian Optimization.For each Gaussian, we quantify the dynamic variation properties of its attributes and classify them into Gaussians with different saliency levels. We then label these Gaussians and optimize the high-saliency ones.}
    \label{fig:frame}
\end{figure*}

To address the spatio-temporal representation limitations of traditional 3D-GS in dynamic scene reconstruction, Wu et al. proposed the 4D Gaussian Splatting (4D-GS)\cite{wu20244d} framework, achieving breakthroughs in dynamic scene modeling through a hierarchical spatio-temporally coupled representation architecture. This method innovatively constructs a hybrid Gaussian deformation field network, whose core consists of a spatio-temporal structure encoder and a lightweight multi-head Gaussian deformation decoder. The former realizes low-rank compressed representation of motion fields through spatio-temporal basis function decomposition, while the latter enables parameterized modeling of cross-frame deformation fields via attention mechanisms.
However, the framework still requires global optimization of millions of Gaussian parameters, where redundant updates of static or low-frequency dynamic parameters significantly degrade convergence efficiency. To resolve this, our study proposes a dynamics-aware parameter update mechanism that adaptively focuses gradient updates based on motion saliency coefficients, thereby accelerating convergence. Furthermore, we introduce a multi-scale manifold enhancement module that synergizes non-linear implicit decoders with explicit deformation fields, enhancing reconstruction capability while preserving spatio-temporal continuity. Our contributions are as follows:\\

\begin{figure*}
    \centering
    \includegraphics[width=1\linewidth]{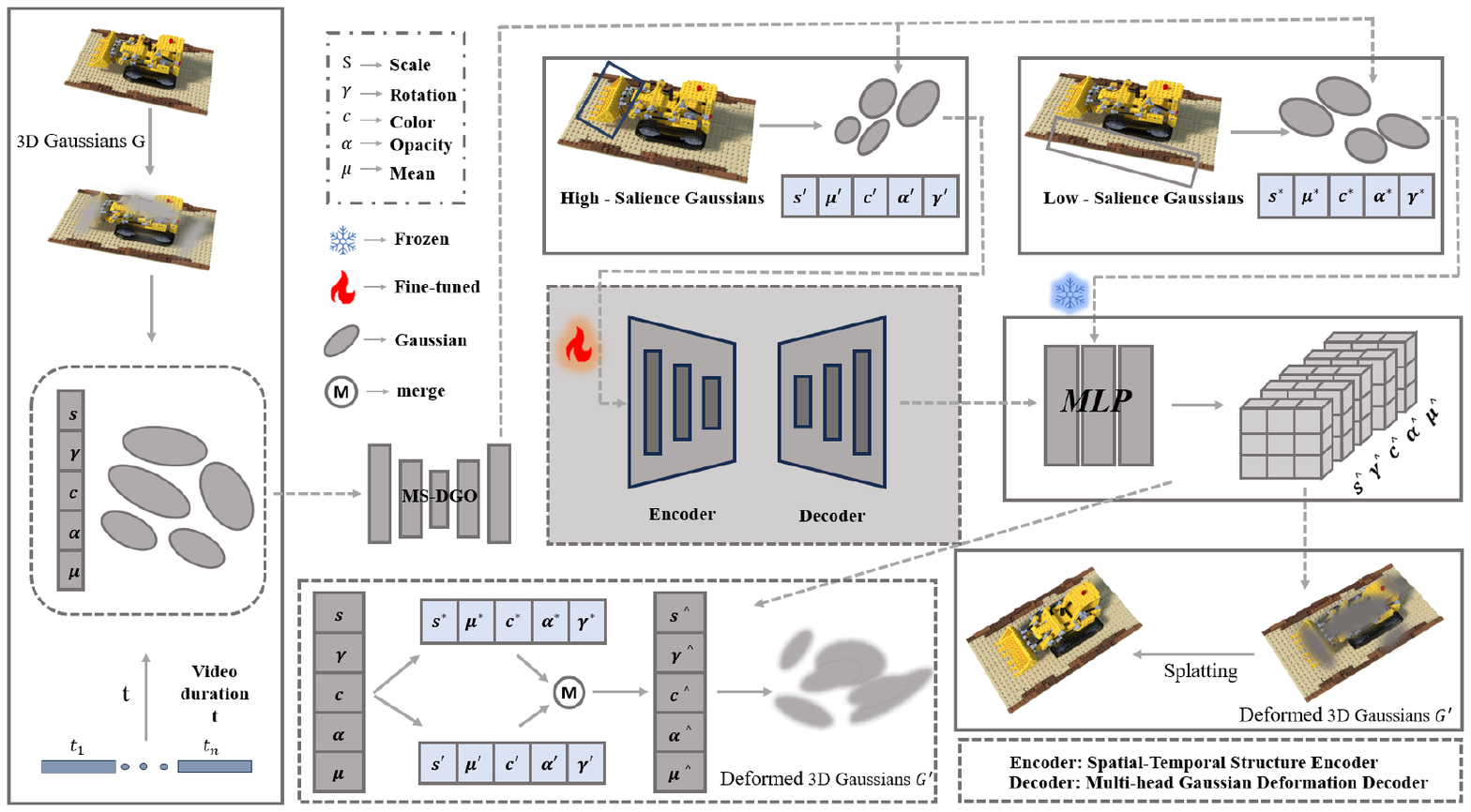}
    \caption{The overall pipeline of our model. For a set of 3D Gaussians $G$, we extract the center coordinates and timestamp $t$ of each Gaussian by querying multi-resolution voxel planes; using $MS-DGO$, we distinguish Gaussians with different saliency coefficients and optimize those with high saliency coefficients; next, we compute the voxel features; decoding these features via a miniaturized multi-head Gaussian deformation decoder yields the deformed 3D Gaussian $G^{\prime}$ at timestamp t; finally, applying Gaussian splatting to the deformed Gaussians generates the final rendered image.}
    \label{fig:frame}
\end{figure*}

$\bullet$ By introducing the Motion Saliency-Driven Dynamic Gaussian Optimization (MS-DGO) method, which quantifies dynamic changes in Gaussian properties to achieve adaptive gradient focusing, distinguishes between Gaussians with varying saliency levels in real time, and implements differentiated optimization, the model's convergence is significantly accelerated while storage redundancy is reduced.\\
$\bullet$ Relying on the synergistic optimization of an implicit nonlinear decoder and an explicit deformation field, it enhances the modeling capability of complex dynamic deformation scenarios, effectively compensates for potential rendering quality degradation caused by MS-DGO, and maintains spatiotemporal continuity.

\section{Related Work}
\subsection{Novel View Synthesis}

Novel view synthesis, as a core technical challenge in 3D reconstruction, has witnessed the emergence of diverse scene representations and rendering strategies in recent years. Among classical explicit-structure methods, approaches like light field mapping \cite{wang2022progressivelyconnectedlightfieldnetwork} , parameterized meshes \cite{yang2022neumeshlearningdisentangledneural,peng20243dmeshnetthreedimensionaldifferentialneural,höllein2022stylemeshstyletransferindoor}, discrete voxels  \cite{zhang2024voxelmambagroupfreestate,han2024mamba3denhancinglocalfeatures,xing2024segmambalongrangesequentialmodeling}, and multiplane projection \cite{liu2025slam3rrealtimedensescene,sun2021neuralreconrealtimecoherent3d}rely on dense supervision for high-fidelity rendering, while NeRF-based methods \cite{kocabas2023hugshumangaussiansplats,barron2022mipnerf360unboundedantialiased} construct continuous scene representations through implicit neural radiance fields, achieving breakthrough progress in view generation accuracy. Addressing dynamic scene modeling demands, researchers in [38,39,42] deconstructed static constraints; particularly, \cite{Fang_2022}  proposed a temporally aware explicit dynamic voxel model, elevating training efficiency to under 30 minutes, a paradigm subsequently advanced in works like \cite{guo2023forward,liu2023robust}. Within dynamic modeling, deformation-driven methods \cite{gao2021dynamic,zhou2023dynpoint} utilize cross-frame optical flow fields for pixel-level spatial transformations, while emerging temporal decoupling neural volumetric techniques \cite{fridovich2023k,gan2023v4d,shao2023tensor4d} significantly accelerate dynamic modeling by decoupling spatial sampling along the temporal dimension. For multi-view systems, studies such as \cite{gao2022mps,wang2023mixed} have designed tailored optimization architectures. However, despite breakthroughs in training efficiency, monocular dynamic scene reconstruction still faces significant real-time inference bottlenecks. This study innovatively constructs a joint optimization framework that, through hierarchical representation design and computational path compression, simultaneously achieves accelerated training and guarantees real-time rendering quality under sparse input conditions.

\subsection{Neural Rendering with Point Clouds}

\section{Preliminary}
This section analyzes the geometric representation and rasterization process of 3D Gaussian Splatting in subsection 3.1.
\subsection{3D Gaussian Splatting}
3D Gaussian distribution is an explicit 3D scene representation that exists in the form of point clouds. Each 3D Gaussian distribution is characterized by a covariance matrix $\Sigma$ and a center point $x$, where $x$ is referred to as the Gaussian mean.
\begin{equation}
G(X) = e^{-\frac{1}{2} x^T \Sigma^{-1} x}.
\end{equation}

To enable separate optimization of parameters, the covariance matrix $\sigma$ can be decomposed into a scaling matrix $S$ and a rotation matrix $R$:
\begin{equation}
\Sigma=\mathbf{RSS}^{T}\mathbf{R}^{T}.
\end{equation}

When rendering a new viewpoint, the system employs differentiable rasterization techniques to project 3D Gaussian distributions onto the camera plane. According to the derivation in literature , the covariance matrix $\Sigma^{\prime}$ in the camera coordinate system can be computed using the following formula, which involves the view transformation matrix $W$ and the Jacobian matrix $J$ of the affine approximation of the projection transformation.
\begin{equation}
\Sigma^{\prime}=JW\Sigma W^TJ^T.
\end{equation}

Each 3D Gaussian contains the following optimizable parameters: a spatial coordinate $X\in R^3$ ,color defined by k-dimensional spherical harmonic coefficients $R\in R^k$
,where $k$ is the order of spherical harmonics.opacity $\alpha\in R$, quaternion rotation parameters $r\in R^4$ and 3D scaling factors $s\in R^3$. During the pixel shading process, the color and opacity values of each Gaussian point are computed based on the radiance field expression in Equation 1.For $N$ ordered Gaussian points covering this pixel, their color blending formula is given by:

\begin{equation}
C=\sum_{i\in N}c_i\alpha_i\prod_{j=1}^{i-1}(1-\alpha_i).
\end{equation}

Here, $c_i$, $\alpha_i$ represents the density and color of this point computed by a 3D Gaussian $G$ with covariance $\Sigma$ multiplied by an optimizable per-point opacity and $SH$ color coefficients.

\section{Method}

\subsection{Motion Saliency-Driven Dynamic Gaussian Optimization}

To accelerate model convergence, this study proposes a dynamically sensitive gradient updating mechanism based on a parametric motion saliency coefficient. This mechanism achieves adaptive gradient focusing by quantifying the dynamic variations in Gaussian attributes. After completing the standard 3D Gaussian splatting framework and deformation field processing, the model performs real-time screening of scene Gaussians according to the motion saliency coefficient, which provides an accurate mathematical representation of Gaussian motion state changes. The screening process employs a preset saliency threshold to partition Gaussians into two categories: high-saliency Gaussians and low-saliency Gaussians.

\begin{figure*}
    \centering
    \includegraphics[width=0.9\linewidth]{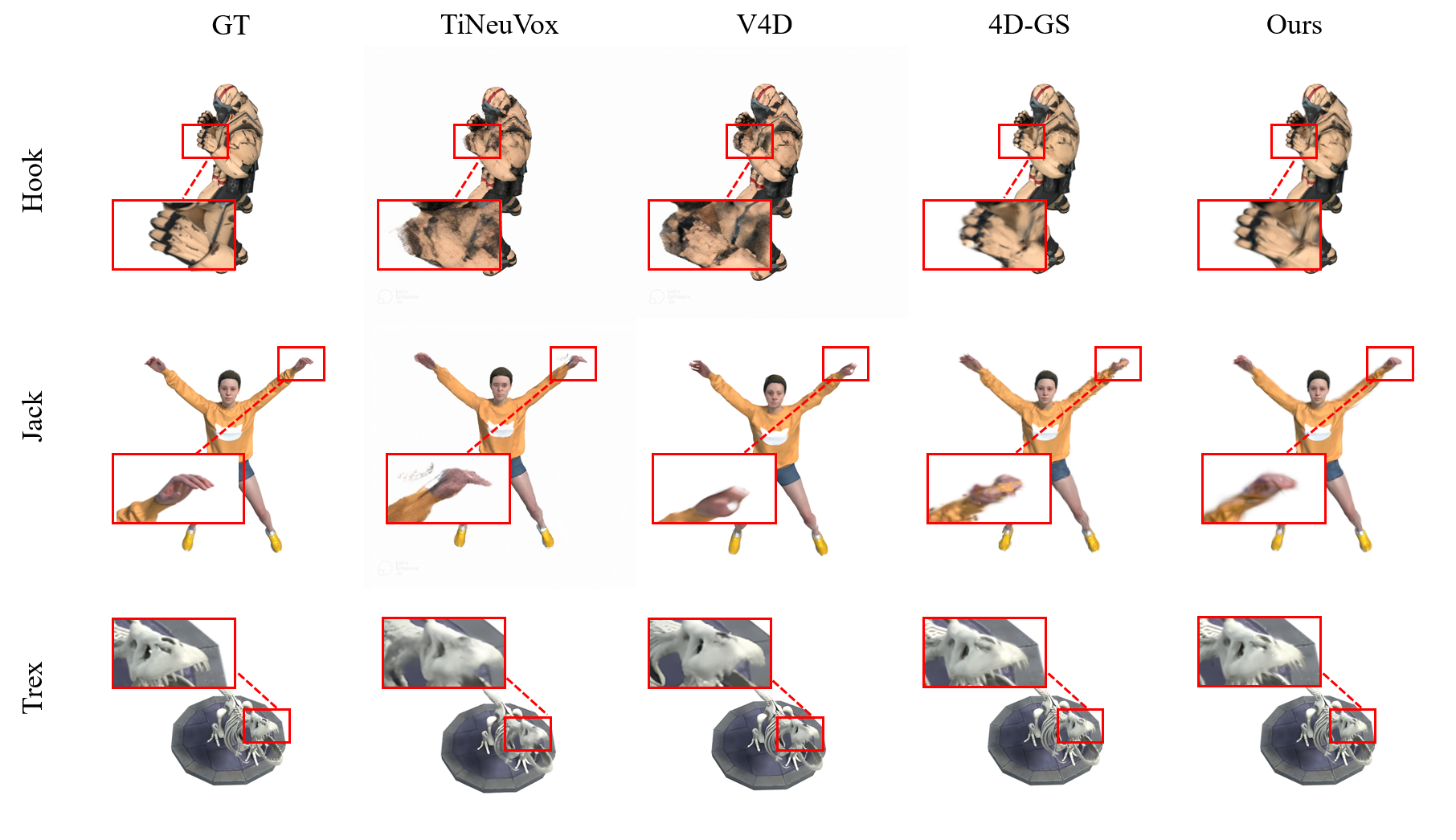}
    \caption{For the synthetic dataset, visualization comparison experiments with other models were conducted \cite{wu20244d,gan2023v4d,fang2022fast}. The rendering results retain the default green background, and this study adopts the rendering parameter configurations of \cite{wu20244d}.}
    \label{fig:frame}
\end{figure*}

High-saliency Gaussians exhibit significant deviation from the target scene state and undergo iterative refinement through deformation fields. Low-saliency Gaussians reach convergence with photorealistic approximation and are exempted from subsequent deformation field updates. By dynamically allocating computational resources and gradient updates, this mechanism continuously focuses optimization on the most promising Gaussian units, substantially enhancing training efficiency while reducing computational redundancy.

\subsection{Multi-scale Disentangled Manifold Deformation}
Upon completion of 3D Gaussian feature encoding, our framework employs a multi-head Gaussian deformation decoder $\mathcal{D} = {(\phi_x, \phi_r, \phi_s)}$ where three independent multilayer perceptrons respectively predict positional deformation $\Delta\mathbf{X} = \phi_x(\mathbf{f}_d)$, rotational deformation $\Delta\mathbf{r} = \phi_r(\mathbf{f}_d)$, and scaling deformation $\Delta\mathbf{s} = \phi_s(\mathbf{f}d)$ using input feature vector $\mathbf{f}d$.To augment reconstruction capacity while strictly preserving spatiotemporal continuity, we introduce a multi-scale manifold enhancement module that achieves hierarchical feature enhancement through synergistic optimization of an implicit nonlinear decoder and explicit deformation field. This module implements three critical operations: dynamic feature dimension adaptation aligning implicit decoder layers with deformation field outputs, cross-scale feature pyramid construction integrating low-frequency geometric structures $\mathcal{L}{\text{c}}$ with high-frequency details $\mathcal{H}$, and nonlinear manifold mapping to boost complex deformation modeling capability.
\begin{equation}
\Delta\mathbf{X}, \Delta\mathbf{r},\Delta\mathbf{s}= \phi(\mathbf{f}_d).
\end{equation}

\begin{equation}
(\mathcal{X^{\prime}},r^{\prime},s^{\prime})=(\mathcal{X}+\Delta\mathcal{X},r+\Delta r,s+\Delta s).
\end{equation}

Finally, we obtain the deformed 3D Gaussians $ G' = \{ x', s', r', \sigma, C \} $.

\section{Experiment}
This section will first present a systematic analysis of the characteristics of the employed datasets, followed by a comparative evaluation of our method's performance across multiple datasets. We then conduct rigorous ablation studies to demonstrate the efficacy of the proposed approach.

\begin{figure*}
    \centering
    \includegraphics[width=1\linewidth]{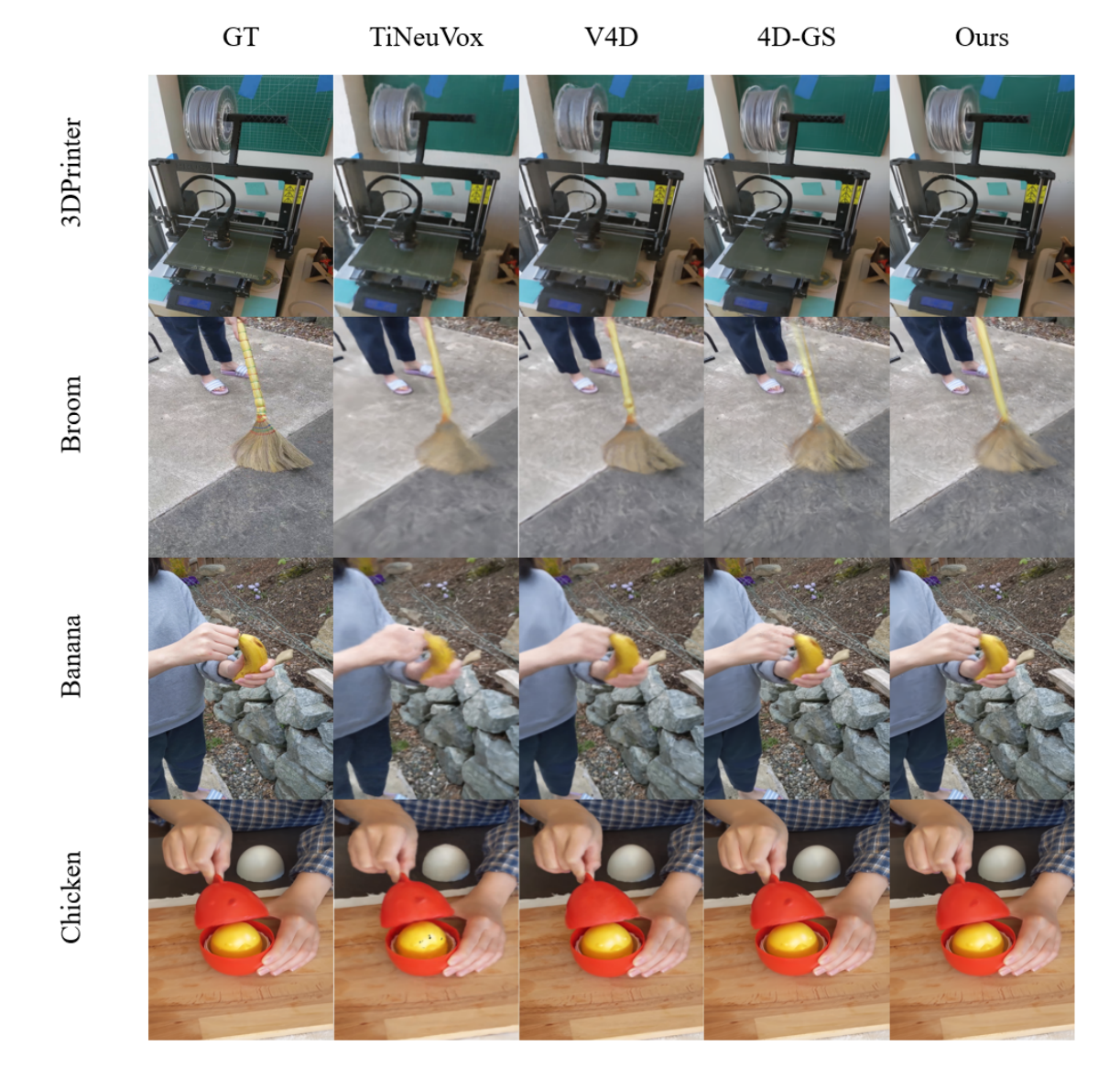}
    \caption{Visualization of the HyperNeRF \cite{park2021hypernerf} dataset compared with other methods\cite{gan2023v4d,fang2022fast,wu20244d} . ‘GT’ stands for ground truth images.Please zoom in for better observation.}
    \label{fig:frame}
\end{figure*}

\subsection{Experimental Settings}
Our implementation is built upon the PyTorch framework, with all experiments conducted on RTX 3090 GPU. We fine-tuned the hyperparameter settings described in 4D-GS and optimized the parameters accordingly.
\subsubsection{Synthetic Dataset.}
This study employs the synthetic dataset constructed by D-NeRF \cite{pumarola2021d} as the primary evaluation benchmark, specifically designed for monocular dynamic scene reconstruction tasks. Its defining characteristics include quasi-randomly distributed camera poses at each temporal node, with dynamic sequence lengths per scene rigorously constrained to the 50-200 frame range.

\subsubsection{Real-world Datasets.}
This study employs the HyperNeRF \cite{park2021hypernerf} dataset  as the benchmark for real-world performance evaluation. The dataset is captured using monocular or binocular cameras, with camera motion constrained to simple linear trajectories. In our experimental pipeline, 200 frames are randomly sampled for modeling analysis, and the initial point clouds are reconstructed via Structure-from-Motion (SfM) algorithm.
\subsubsection{Dynamic Object dataset.}
We also used another artificially synthesized dynamic 3D dataset, namely the Dynamic Object dataset \cite{li2023nvfi}. This dataset contains 6 different 3D objects, each exhibiting unique movement patterns, including rigid or deformable motions in 3D space. Examples include Bat and Fan, among others.

\begin{figure*}
    \centering
    \includegraphics[width=1\linewidth]{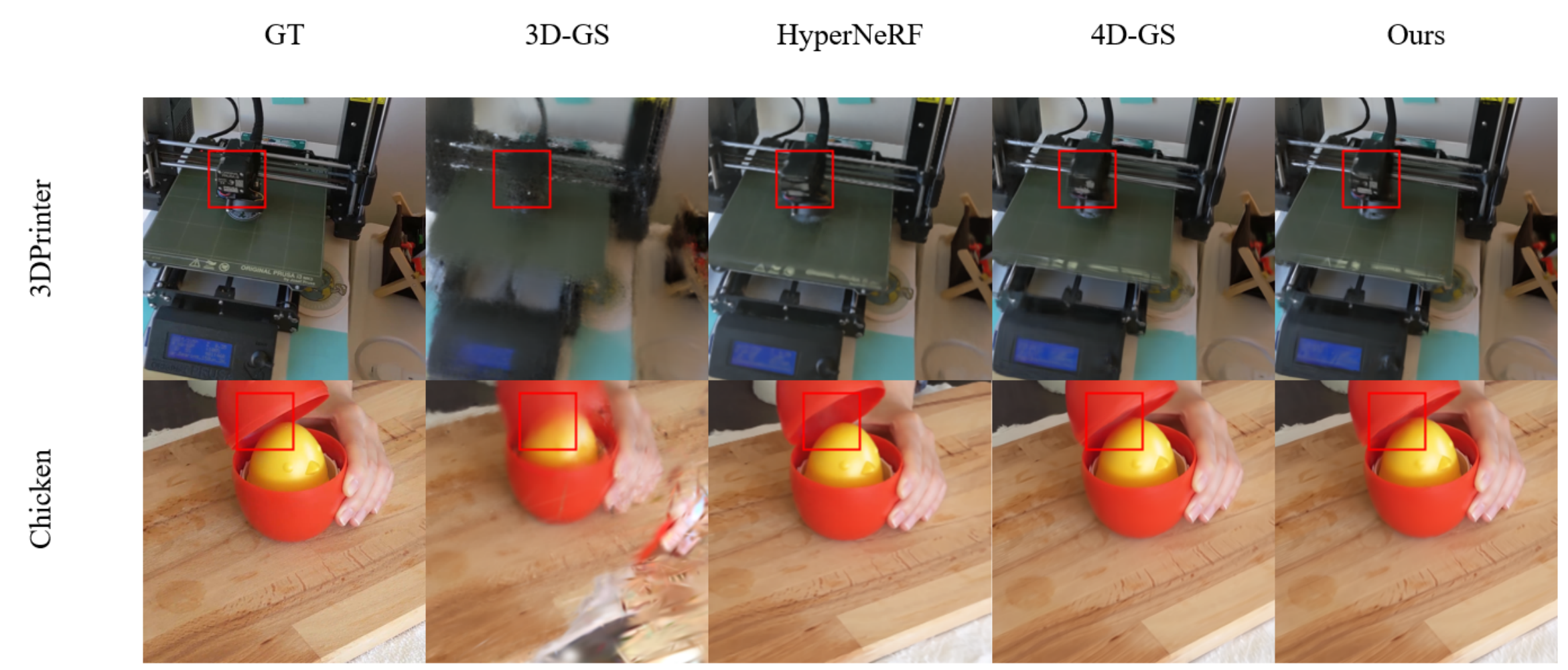}
    \caption{Experimental results show that in the visualization results, our method can recover more details compared to other methods \cite{kerbl20233d,park2021hypernerf,wu20244d}.}
    \label{fig:frame}
\end{figure*}

\begin{table}[htbp]
\centering
\caption{Quantitative results on the synthetic dataset. The best and the second best results are denoted by pink and yellow. The rendering resolution is set to 800×800. “Time” in the table stands for training times.}
\label{tab:model_performance}
\setlength{\tabcolsep}{5pt}  
\begin{tabular}{l|c|c|c|c|c}
\toprule
Model & PSNR (dB) & SSIM & LPIPS & Time & FPS \\
\midrule
TiNeuVox-B {\cite{fang2022fast}}    & 32.67 & 0.97 & 0.04 & 28 mins   & 1.5 \\
KPlanes {\cite{fridovich2023k}}      & 31.61 & 0.97 & -    & 52 mins   & 0.97 \\
HexPlane-Slim {\cite{cao2023hexplane}} & 31.04 & 0.97 & 0.04 & 11m 30s   & 2.5 \\
3D-GS {\cite{kerbl20233d}}       & 23.19 & 0.93 & 0.08 & 10 mins   & \cellcolor{pink!80}170 \\
FFDNeRF {\cite{guo2023forward}}  & 32.68 & 0.97 & 0.04 & -         & $<1$ \\
MSTH {\cite{wang2023masked}}     & 31.34 & 0.98 & 0.02 & \cellcolor{pink!80}6 mins    & - \\
V4D {\cite{gan2023v4d}}          & 33.72 & 0.98 & 0.02 & \cellcolor{yellow!60}6.9 hours & 2.08 \\
4D-GS {\cite{wu20244d}}          & \cellcolor{yellow!60}34.05 & \cellcolor{yellow!60}0.98 & \cellcolor{yellow!60}0.02 & 8 mins     & 82 \\
Ours           & \cellcolor{pink!80}34.56 & \cellcolor{pink!80}0.98 & \cellcolor{pink!80}0.02 & 8 mins     & \cellcolor{yellow!60}96 \\
\bottomrule
\end{tabular}
\end{table}

\subsection{Results}
This study conducts a systematic evaluation of the experimental results based on a multi-dimensional metric system, specifically covering core evaluation criteria such as peak signal-to-noise ratio (PSNR), perceptual quality metric LPIPS, and structural similarity index (SSIM). To address the requirements for visual quality assessment in the novel view synthesis task, comparative experiments are performed with representative state-of-the-art methods in the field, including the methods proposed in literatures \cite{fang2022fast,fridovich2023k,cao2023hexplane,kerbl20233d,guo2023forward,wang2023masked,gan2023v4d,wu20244d,park2021nerfies,park2021hypernerf}. 
Among these, the experimental results of other comparative methods on the synthetic dataset are directly cited from the original literature of 4D-GS. The quantitative analysis results on the synthetic dataset are detailed in Table 1. Notably, although existing dynamic hybrid representation methods can achieve relatively high-quality reconstruction effects, limitations persist in the optimization of their dynamic motion modeling components, leaving room for improvement in the detail reconstruction performance of methods such as 4D-GS. In sharp contrast, our method not only achieves optimal rendering quality on the synthetic dataset but also exhibits superior convergence efficiency while maintaining an extremely low level of storage consumption.

\begin{table}[htbp]
\centering
\caption{Quantitative results on HyperNeRF [39] vrig dataset with the rendering resolution of 960×540.}
\label{tab:model_performance_no_color}
\setlength{\tabcolsep}{5pt}  
\begin{tabular}{l|c|c|c|c}
\toprule
Model & PSNR (dB) & MS-SSIM & Times & FPS \\
\midrule
Nerfies {\cite{park2021nerfies}}    & 22.2 & 0.803 & $\sim$ hours & $<1$ \\
HyperNeRF {\cite{park2021hypernerf}}  & 22.4 & 0.814 & 32 hours     & $<1$ \\
TiNeuVox-B {\cite{fang2022fast}}  & 24.3 & 0.836 & 30 mins      & 1    \\
3D-GS {\cite{kerbl20233d}}      & 19.7 & 0.680 & 40 mins      & \cellcolor{pink!80}55   \\
FFDNeRF {\cite{guo2023forward}}    & 24.2 & 0.842 & -            & 0.05 \\
V4D {\cite{gan2023v4d}}        & 24.8 & 0.832 & 5.5 hours    & 0.29 \\
4D-GS {\cite{wu20244d}}                & \cellcolor{yellow!60}25.2 & \cellcolor{yellow!60}0.845 & \cellcolor{yellow!60}30 mins     & \cellcolor{yellow!60}34   \\
Ours                & \cellcolor{pink!80}25.5 & \cellcolor{pink!80}0.853 & \cellcolor{pink!80}26 mins      & 28   \\
\bottomrule
\end{tabular}
\end{table}


The experimental results on the real-scene dataset are detailed in Table 2. Notably, our method exhibits significant advantages in comprehensive performance compared to partial NeRF methods and other grid-based neural radiance field methods \cite{fang2022fast,cao2023hexplane,fridovich2023k,wang2023masked}.. Further comparison with the 4D-GS method reveals that our method demonstrates a slight leading edge in its core evaluation metrics. From an overall performance perspective, our method is on par with existing mainstream methods in terms of rendering quality, while boasting more efficient convergence characteristics and exhibiting superior real-time performance in indoor scene free-view rendering tasks.

\subsection{Ablation experiment}
\subsubsection{Motion Saliency-Driven Dynamic Gaussian Optimization.}
The MS-DGO mechanism achieves adaptive gradient focus by quantifying the dynamic changes of Gaussian attributes, which can be interpreted through $\phi_r$. Dynamically partitions scene Gaussians into two categories in real time: high-saliency Gaussians and low-saliency Gaussians. For high-saliency Gaussians, iterative optimization is performed using deformation fields; low-saliency Gaussians, upon meeting the convergence criteria of approximating the real state, halt subsequent deformation field updates. Experimental results demonstrate that by dynamically allocating computational resources and optimizing focus, this mechanism effectively accelerates model convergence and reduces storage requirements. When this module is removed, and only deformation fields are employed to continuously optimize a large number of low-saliency Gaussians, experiments observe a significant increase in storage overhead and a notable slowdown in convergence speed.

\begin{table}[htbp]
\centering
\caption{Ablation studies conducted using our proposed method on the Dynamic Object dataset.}
\label{tab:ablation_studies_filtered}
\setlength{\tabcolsep}{5pt}  
\begin{tabular}{l|ccc|c|c}
\toprule
Model & PSNR(dB) & SSIM & LPIPS & Time & FPS  \\
\midrule
Ours w/o $\phi_r$ & 24.429 &  \cellcolor{yellow!60}0.950 & 0.058 & \cellcolor{pink!80}12.5 mins & \cellcolor{pink!80}72  \\
Ours w/o $\phi_s$ &  \cellcolor{yellow!60}24.518 & 0.949 &  \cellcolor{yellow!60}0.058 & 14 mins & 63  \\
Ours              & \cellcolor{pink!80}24.534 & \cellcolor{pink!80}0.952 & \cellcolor{pink!80}0.059 & \cellcolor{yellow!60}13.5 mins & \cellcolor{yellow!60}66 \\
\bottomrule
\end{tabular}
\end{table}

\begin{figure*}
    \centering
    \includegraphics[width=0.9\linewidth]{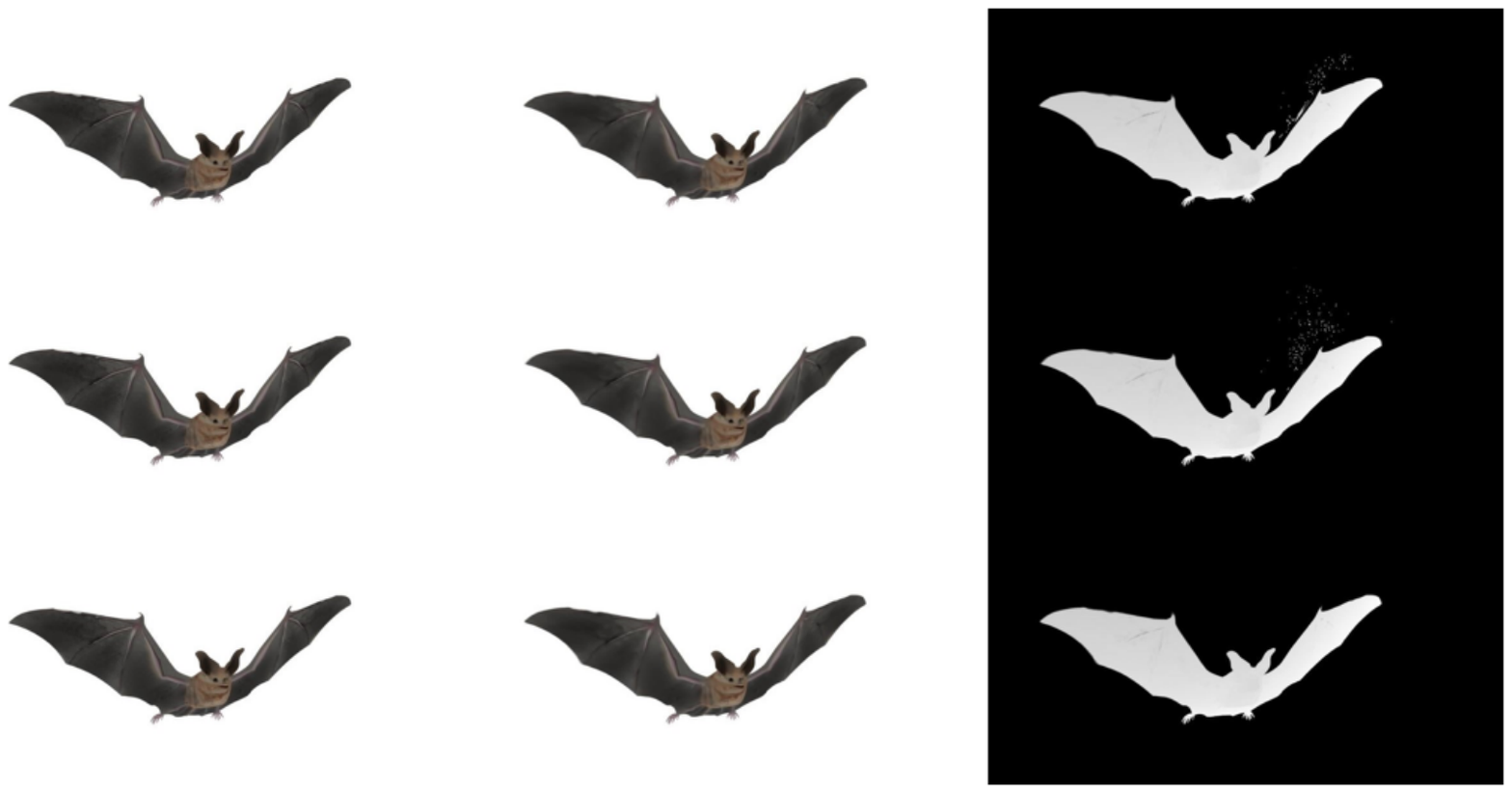}
    \caption{Visualization of tracking with 3D Gaussians.From top to bottom are the quantitative visualization results of Ours w/o $\phi_r$, Ours w/o $\phi_s$, and Ours, respectively.}
    \label{fig:frame}
\end{figure*}

\subsubsection{Multi-scale Disentangled Manifold Deformation.}
\begin{figure*}
    \centering
    \includegraphics[width=1\linewidth]{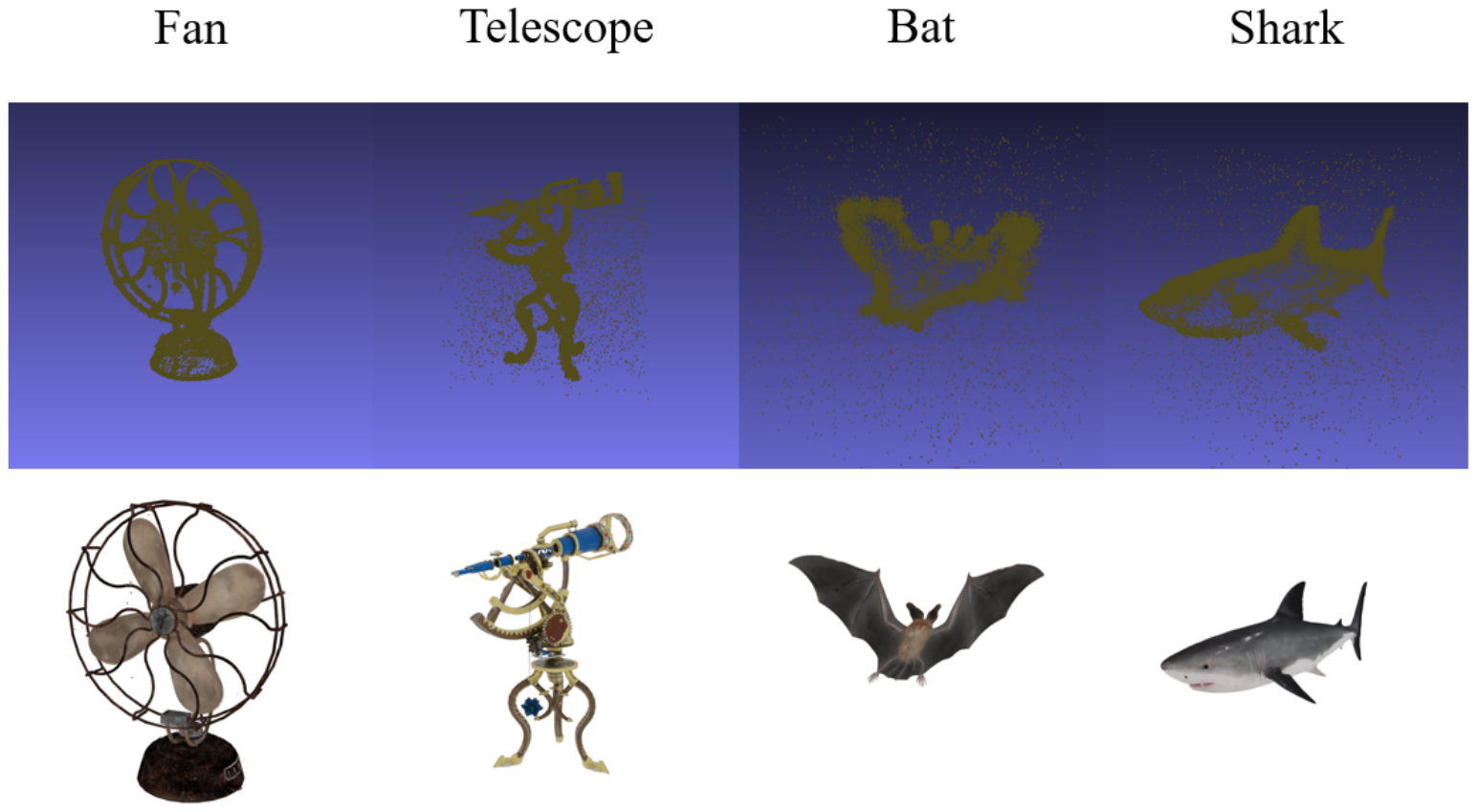}
    \caption{Point cloud distribution visualization.}
    \label{fig:frame}
\end{figure*}
To further enhance reconstruction quality and strictly maintain spatiotemporal continuity, this study introduces a multi-scale manifold enhancement module that can be interpreted via $\phi_s$. This module strengthens the modeling capability of complex deformation scenarios by leveraging a cooperative optimization mechanism between an implicit nonlinear decoder and an explicit deformation field, combined with a nonlinear manifold mapping technique. Experimental results show that removing this module leads to a significant degradation in rendering quality. Although the MS-DGO mechanism introduces a certain degree of rendering quality loss, the incorporation of this module partially offsets such impacts, thereby balancing the overall performance.

\section{Conclusion}
This paper addresses the critical challenge of balancing model convergence speed and rendering quality in 3D dynamic scene reconstruction by proposing a method named GS-DMSR. Through quantifying the dynamic changes of Gaussian attributes to achieve adaptive gradient focusing, along with synergistic optimization between an implicit nonlinear decoder and an explicit deformation field, it significantly accelerates convergence, reduces storage redundancy, and strengthens the modeling capability of complex deformation scenarios. This approach effectively achieves a balance between efficient convergence and real-time rendering.

\bibliographystyle{ieeetr}
\bibliography{reference}
\end{document}